\documentclass[preprint,12pt]{elsarticle}

\usepackage{amssymb}
\usepackage{amsmath}
\usepackage{amsfonts}
\usepackage{algorithmic}
\usepackage{graphicx}
\usepackage{multirow}
\usepackage{float}
\usepackage{subcaption}
\usepackage{textcomp}
\usepackage{xcolor}
\usepackage{array}
\usepackage{tabularx}
\usepackage{comment}

\journal{Engineering Applications of Artificial Intelligence}

\begin{document}

\begin{frontmatter}



\title{Optimal Sensor Placement in Power Transformers Using Physics-Informed Neural Networks}

 %% Author name


\author[1]{Sirui Li}
\affiliation[1]{organization={Division of Decision and Control Systems, Department of Intelligent Systems, KTH Royal Institute of Technology}, 
  city={Stockholm},
  country={Sweden}}
%\authornote{}
%\email{bragone@kth.se}

\author[2]{Federica Bragone}
\affiliation[2]{organization={Division of Computational Science and Technology, Department of Computer Science, KTH Royal Institute of Technology }, 
  city={Stockholm},
  country={Sweden}}

\author[1]{Matthieu Barreau}
%\affiliation[3]{organization={Division of Decision and Control Systems, KTH Royal Institute of Technology }, 
%  city={Stockholm},
%  country={Sweden}}

\author[3]{Tor Laneryd}
\affiliation[3]{organization={Hitachi Energy Research}, 
  city={Västerås},
  country={Sweden}}

\author[1]{Kateryna Morozovska}

%\author[1]{Stefano Markidis}

%% Abstract
\begin{abstract}
Our work aims at simulating and predicting the temperature conditions inside a power transformer using Physics-Informed Neural Networks (PINNs). The predictions obtained are then used to determine the optimal placement for temperature sensors inside the transformer under the constraint of a limited number of sensors, enabling efficient performance monitoring. The method consists of combining PINNs with Mixed Integer Optimization Programming to obtain the optimal temperature reconstruction inside the transformer. 
%First, we introduce a novel approach for the regularization of the PINN model through physical unit scaling. 
First, we extend our PINN model for thermal modelling of power transformers to solve the heat diffusion equation from 1D to 2D space. Finally, we construct an optimal sensor placement model inside the transformer that can be applied to problems in 1D and 2D.
\end{abstract}



%% Keywords

\end{frontmatter}



\section{Introduction}
Temperature monitoring of power components is an important factor to ensure their longevity and optimize operation and maintenance needs. Traditional methods for temperature evaluation are often limited by long computing times and a larger need for memory to solve the problem numerically. Therefore, researchers have been looking into using Physics-Informed Neural Networks (PINNs) \cite{RAISSI} for estimating the thermal performance of the power components. Most of these works are oriented towards power transformers as they play a major role in power distribution \cite{Bragone1649301, Bragone1803600, Fede3}. \\
Earlier works with the application of PINN in the energy domain are often related to applications in areas like power systems, high voltage components, aging estimation, and heat exchangers. For example, authors of works \cite{Spyros1, Spyros2, Spyros4} and \cite{Spyros3} explore how power systems engineers can benefit from using PINNs and deep learning to optimize power delivery and system performance. A few works explore the benefits of using PINNs for lifetime estimation of renewable power plants \cite{Joxe1}, power transformers \cite{Joxe2}, and even electrical insulation \cite{Fede4}. However, there is a lack of existing general knowledge on using PINNs for decision-making and control of energy systems and individual components, with only local solutions presented for specific problems like system identification \cite{Spyros5}. Therefore, we aim to find a generalizable strategy for integrating PINNs into the decision-making process in energy engineering. The presented study focuses on temperature monitoring in power transformers, which can be extended further for similar problems in energy engineering.\\
Temperature monitoring is a key factor for the safe operation and maintenance of power transformers. The common methods for thermal analysis use computational fluid dynamics to model, such as the mineral-oil-immersed transformer windings \cite{IEC60076-7, IEEE} and the thermal circuit of the transformer \cite{susa2005dynamic, susa2006dynamic, susa2006dynamic2}. These models allow the calculation of the critical temperature and can also be used to simulate and analyze the internal temperature of the transformer given predefined weather and load conditions. However, the computational complexity of numerical methods increases exponentially with the complexity of the model, and its accuracy depends on the suitability of the model grid discretization. Therefore, in order to reduce the model complexity and adapt transformer temperature monitoring to real-time decision-making, data-driven methods, specifically PINNs, are explored in more detail in \cite{Fede3, Fede2, Fede1}. While proposed PINN models allow the simulation and prediction of the internal temperature of the transformer to be faster than numerical methods, they still require substantial computational resources and time to converge. \\
In our work, we aim to address the limitation of computational complexity when solving heat diffusion problems in power transformers with PINNs by introducing new temperature sensors. By adding additional data points inside the domain, we could reduce the problem's size and ensure faster response times to temperature changes. In order to find the best location and the optimal number of sensors for temperature monitoring, we integrate PINN solution into a mixed integer linear programming (MILP) model. On this basis, we introduce a novel approach to decision-making and data collection in power components by integrating the PINN solution at the component design stage to determine the minimum number of most stable high-temperature points, which can effectively serve as a basis for faster and reliable real-time solution of the transformer thermal model. %fextended and applied this method to find the most suitable internal temperature monitoring positions inside of the transformer under the restriction of a limited number of temperature sensors to achieve efficient performance monitoring. 
In addition to the 1D spacial model from \cite{Fede3}, we extend the analysis to the 2D spacial model to validate the results for the sensor placement solution.

\section{Methods}
This section introduces the methods used for the study, including a description of the model for the temperature distribution in power transformers. We also introduce the PINN structure in 1D and 2D and describe the three proposed optimization models for the optimal sensor location for temperature detection.

\subsection{Heat Diffusion Problem}
\begin{figure}[H]
     \centering
     \begin{subfigure}[b]{0.49\textwidth}
         \centering
         \includegraphics[width=\textwidth]{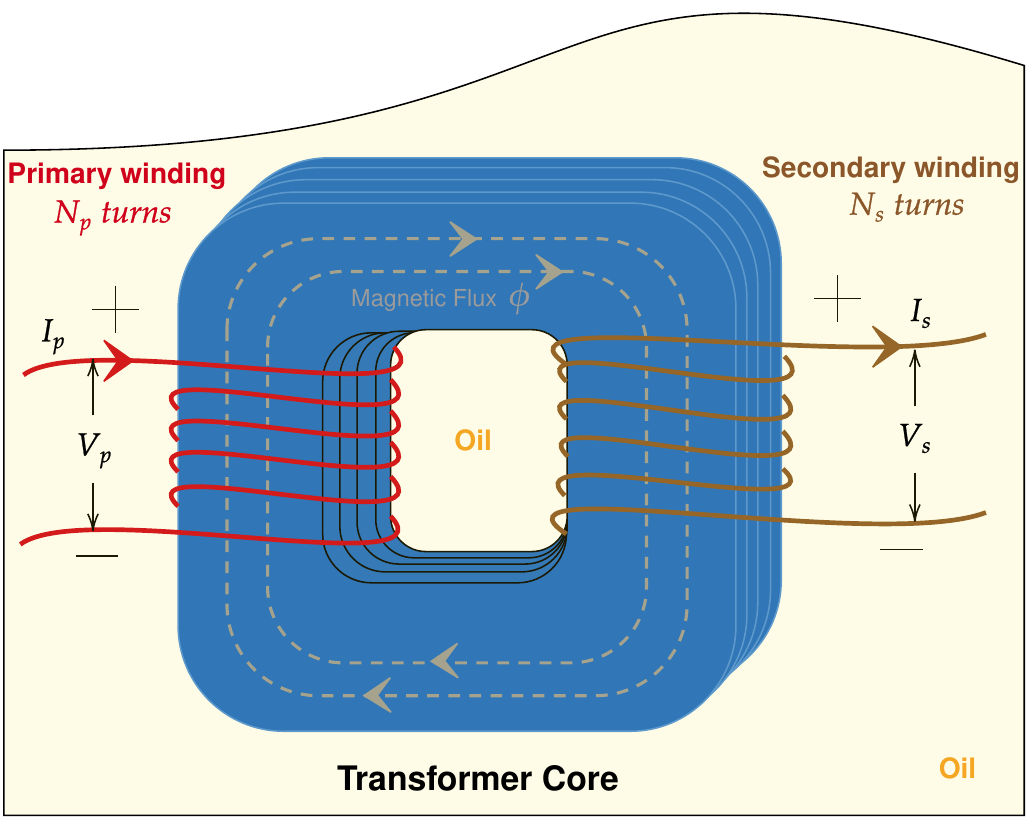}
         \caption{Example of a simple transformer.}
         \label{fig:orginal_Tansformer}
     \end{subfigure}
     \hfill
     \begin{subfigure}[b]{0.49\textwidth}
         \centering
         \includegraphics[width=\textwidth]{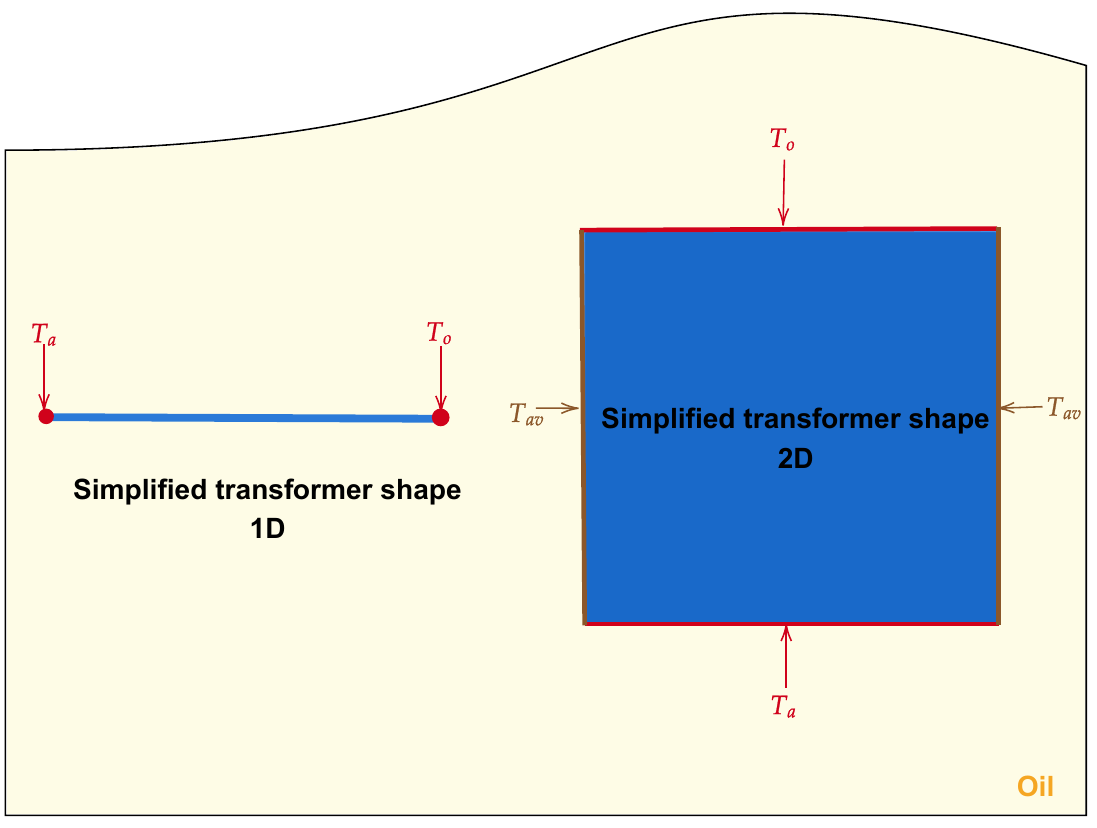}
         \caption{Graphically simplified transformer shape.}
         \label{fig:Simplied_transformer}
     \end{subfigure}
\caption{On the left, a transformer placed in oil, where $I_p$, $I_s$, $V_p$, and $V_s$ are the primary current, secondary current, primary voltage, and secondary voltage, respectively. On the right, simplification of the transformer structure for models in 1D and 2D.}
\label{fig: Tansformer}
\end{figure}

The shape of the actual power transformer is relatively complex. For convenience of study, we have simplified its shape. In previous studies, the transformer was described as a line $\mathbf{x} = (x) \in [0, 1]$, and it was assumed that it was immersed in oil as a coolant. When the spatial dimension of the model is extended to 2D space $\mathbf{x} = (x,y) \in [0, 1]^2$, it is assumed that the coolant remains unchanged, and the transformer shape is described as a square, see Figure \ref{fig: Tansformer}. We define $\Omega$ as the space domain, $\Omega = [0, 1]$ in 1D and $\Omega = [0, 1]^2$ in 2D.

The general form of the heat diffusion equation for the 1D and 2D model is given by: 
\begin{equation}
    \rho c_p \frac{\partial u}{\partial t}  = k \Delta_{\mathbf{x}} u + q
\end{equation}
where $\rho$ is the density, $c_p$ is the heat capacity, $k$ is the thermal conductivity, $\Delta_{\mathbf{x}}$ is the Laplace operator. The term $q$ represents the heat source that, for this problem, we define as:
 \begin{align}
     & q = q(\mathbf{x},t) = (P_0 + P_K(\mathbf{x},t) - h(u(\mathbf{x},t) - T_a(t))),\\
     & P_K(\mathbf{x},t) = P_K^t(t)P_K^\mathbf{x}(\mathbf{x}),
 \end{align}
 where $P_0$ is the no-load loss, $P_K(\mathbf{x},t)$ is the load loss, $h$ is the convective heat transfer coefficient, and $T_a(t)$ is the ambient temperature. The load loss has a component dependent on time, $P_K^t(t)$, and one depending on space, $P_K^{\mathbf{x}}(\mathbf{x})$, which differ for the 1D and 2D problems. Their forms are given by:

 \begin{align}
   % \begin{array}{l}
        & P_K^t(t) = \nu K(t)^2  \\
        & P_K^{\mathbf{x}}(\mathbf{x}) = \begin{cases}
             0.5\sin(3 \pi x) + 0.5, & \text{ if } \Omega = [0, 1], \\
             1, & \text{ if } \Omega = [0, 1]^2.
         \end{cases} 
    %\end{array}
\end{align}
 where $K(t)$ is the load factor, and $\nu$ is the rated load loss. The boundary conditions for the 1D problem are defined as:
 \begin{align}
    & u(0,t) = T_a, \\
    & u(1,t) = T_o,
\end{align}

while for the 2D problem are:
 \begin{align}
    & u(0,y,t) = T_a, \\
    & u(1,y,t) = T_o, \\
    & u(x,0,t) = u(x,1,t) = \frac{T_a + T_o}{2} = T_{av}
\end{align}
 where $T_a$ is the ambient temperature, $T_o$ is the top oil temperature, and we define $T_{av}$ The average temperature between $T_a$ and $T_o$.
 Table \ref{tab:phys_params_eq} shows the values used for the parameters of the heat diffusion equation in 1D and 2D. 

\begin{table}[H]
\renewcommand{\arraystretch}{1.2}
\centering
\caption{Physical parameters and corresponding values of the heat diffusion equation in 1D and 2D.}
\label{tab:phys_params_eq}
\begin{tabular}{|l|l|m{4em}|m{4em}|}
\hline
\textbf{Parameters} & \textbf{Unit}  & \textbf{1D}  & \textbf{2D} \\ \hline
Thermal conductivity, $k$ & [$W/m\cdot K$] &  \multicolumn{2}{c|}{$50$}\\ \hline
Density, $\rho$ & [$kg/m^3$] &  \multicolumn{2}{c|}{$900$}  \\ \hline
Heat Capacity, $c_p$  & [$J/kg\cdot K$]  &  \multicolumn{2}{c|}{$2000$}     \\ \hline
Heat Transfer Coefficient, $h$  & [$W/m^2\cdot K$] & 1000 & 2000                 \\ \hline
No-Load Loss, $P_0$  &  [$W$] &  \multicolumn{2}{c|}{$1500$} \\ \hline
Rated Load Loss, $\nu$ & [$W$]  &  \multicolumn{2}{c|}{$83000$} \\ \hline
\end{tabular}
\end{table}

\begin{figure}[ht]
     \centering
     \includegraphics[width=0.7\textwidth]{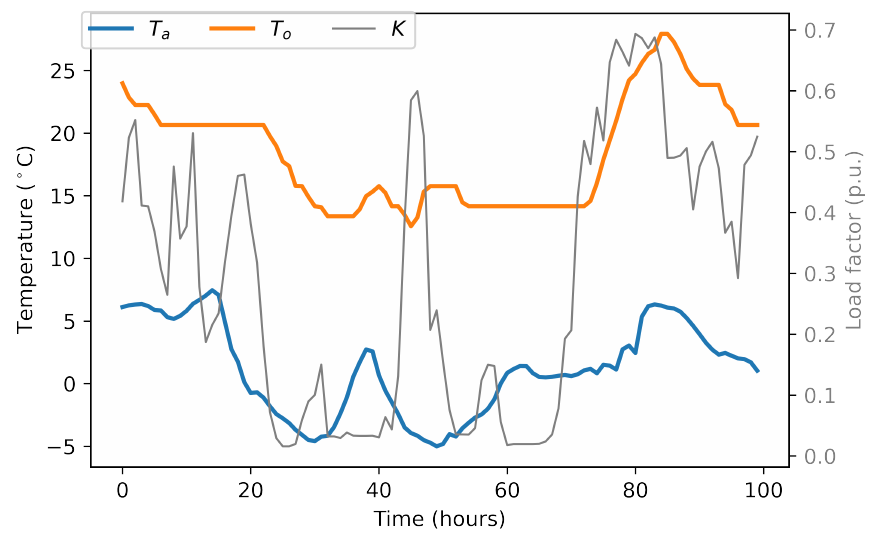}
    \caption{Ambient temperature, top oil temperature, and load factor measurements during the first $100$ hours.}
    \label{fig:input_data}
\end{figure}

Figure \ref{fig:input_data} represents the data used for the problem. In particular, it consists of the ambient temperature $T_a$ [$^\circ $C], the top oil temperature $T_o$ [$^\circ$C], and the load factor $K$ [p.u.], corresponding to the red, blue, and grey lines in the plot for the first 100 hours, that we are considering, of the dataset.

\subsection{PINNs}

\begin{figure}[ht]
    \centering
         \includegraphics[width=0.9\textwidth]{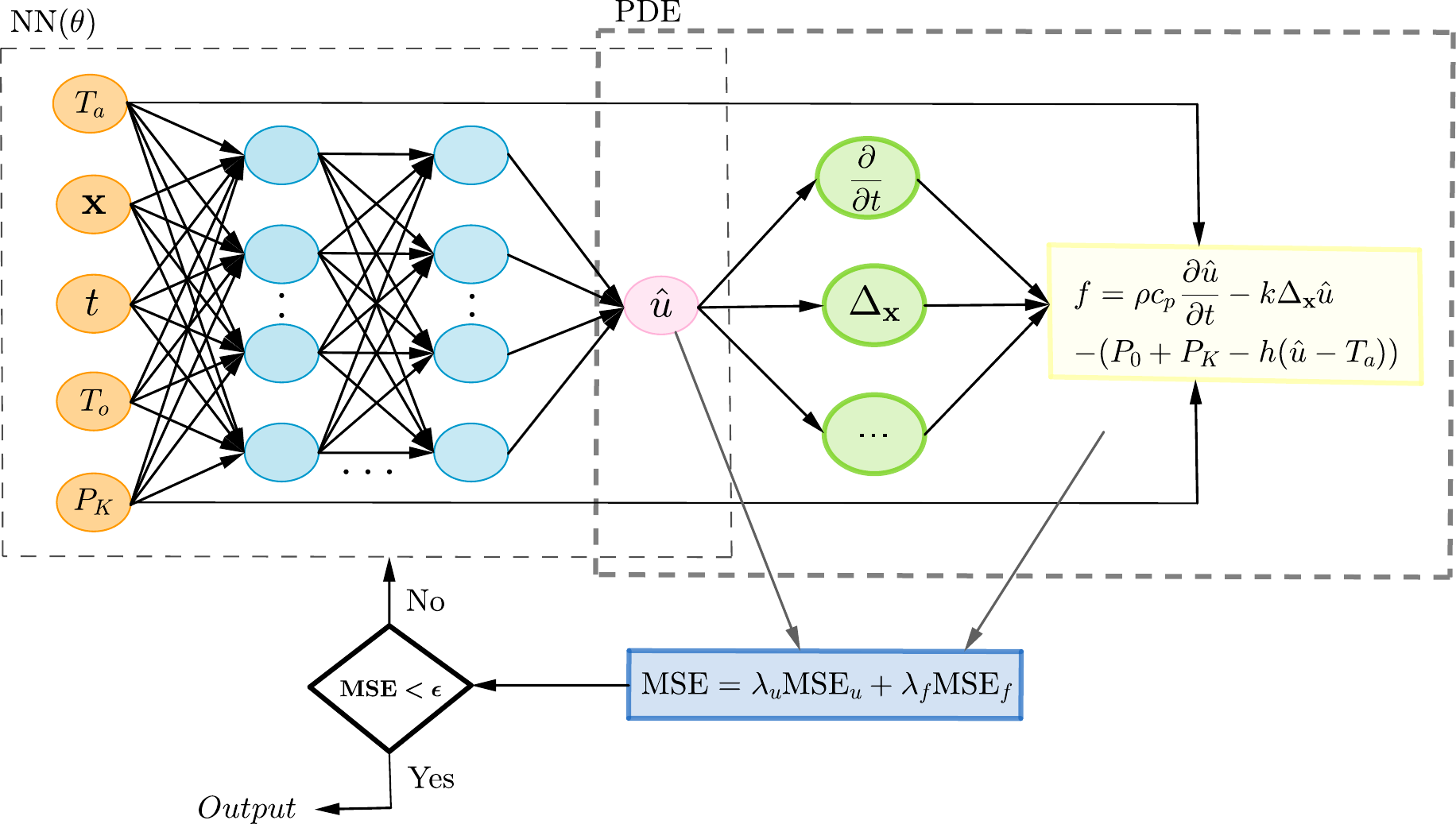}
         \caption{Structure of the PINN model.}
\label{fig:PINN}
\end{figure}

The structure of the PINN model is shown in Figure \ref{fig:PINN}, consisting of a neural network part approximating the solution $u$ from the inputted values and a residual side where the partial derivatives of the considered equation are evaluated using automatic differentiation \cite{baydin2018automatic}.

We define the residual $f$ to be our heat diffusion equation:
\begin{equation}
\label{eq:residual_function}
    f(\mathbf{x},t) =  \rho c_p\frac{\partial u}{\partial t}  - k\Delta_{\mathbf{x}} u - (P_0 + P_K(\mathbf{x},t) - h(u(\mathbf{x},t) - T_a(t))).
\end{equation}

The overall loss function of the PINN model is defined as the weighted sum of the mean-squared error assigned to the boundary conditions, MSE$_u$ and the mean squared error of the residual, MSE$_f$:
\begin{equation}
    MSE = \lambda_u MSE_u + \lambda_f MSE_f,
\end{equation}
where
\begin{align}
    & MSE_u = \frac{1}{N_u} \sum_{i=1}^{N_u} |\hat{u}(\mathbf{x}_u^i, t_u^i)-u^i|^2, \label{eq:mse_u} \\
    & MSE_f =  \frac{1}{N_f} \sum_{i=1}^{N_f} |f(\mathbf{x}_f^i, t_f^i)|^2.  \label{eq:mse_f}
\end{align}
From the equations, $\{\mathbf{x}_u^i, t_u^i, u^i\}_{i=1}^{N_u}$ corresponds to the training data for the boundary conditions; $\hat{u}$ is the approximation of the solution $u$ at the training boundary coordinates $\mathbf{x}_u$ and $t_u$; $\{\mathbf{x}_f^i, t_f^i\}_{i=1}^{N_f}$ are the collocation points of the residual $f$; $N_u$ is the number of boundary training points; $N_f$ are the number of collocation points; $\{\lambda_u, \lambda_f\}$ are the weights assigned to the corresponding MSE.

The structure of the PINN model consists of one input layer with $4+\delta$ neurons, where $\delta$ corresponds to the spatial dimension of the problem, which is either 1 or 2 in this case. There are four hidden layers with 50 neurons each and one output layer with one neuron corresponding to the solution $u$. The input values are standardized to ensure the model's stability and efficiency during training according to \cite{shanker1996effect}. Then, output values are normalized.
%Furthermore, the physical regularization part of the loss function is scaled to reduce the poor model training effect caused by the large difference between the data. Considering that the problem is a transient heat transfer problem, and the SI unit of time is the only unit that is not scaled by $10$ times, we choose to strictly scale according to the SI unit of the parameters and to convert the time units from hours to seconds. Such scaling is defined as physical scaling. This scaling is defined as physical scaling while scaling of all parameters by a fixed factor $\beta$ in previous studies \cite{Fede1, Fede2, Fede3}, is defined as mathematical scaling. Other hyperparameters used in the PINN model are defined in Table %\ref{tab:all hyper_param}, \ref{tab:1D hyper_param}.
Furthermore, the parameters of the residual function are scaled using a fixed factor $\beta=1000$ introduced in previous studies using the same model \cite{Fede1, Fede2, Fede3}. Other hyperparameters used in the PINN model are defined in Table \ref{tab:1D_hyper_param}.

\begin{table}[ht]
\renewcommand{\arraystretch}{1.2}
\centering
\caption{Hyperparameters for 1D and 2D PINN models}
\begin{tabular}{|l|m{4em}|m{4em}|}
\hline
\textbf{Parameter}                                            & \textbf{1D} & \textbf{2D} \\ \hline
Number of hidden layers                & \multicolumn{2}{c|}{$4 $}                  \\ \hline
Number of neurons of the hidden layers & \multicolumn{2}{c|}{$50$}                  \\ \hline
Number of neurons of the input layer   & \multicolumn{2}{c|}{space dimension $+ 4$} \\ \hline
Number of neurons of the output layer  & \multicolumn{2}{c|}{$1 $}                  \\ \hline
Activation function                            & \multicolumn{2}{c|}{tanh}                \\ \hline
Weight initialization                          & \multicolumn{2}{c|}{Xavier}              \\ \hline
Optimizer                                      & \multicolumn{2}{c|}{Adam, L-BFGS-B }     \\ \hline
Epochs per training, {[}Adam, L-BFGS-B{]}      & \multicolumn{2}{c|}{$[5000, 5000]  $}        \\ \hline

Adam learning rate                                            & $1e-6 $   &  $1e-4 $        \\ \hline
Adam epsilon                                                  & \multicolumn{2}{c|}{ $1e-5 $ }          \\ \hline
%Number of epochs Adam per iteration & \multicolumn{2}{c|}{$5000 $}        \\ \hline
%Number of epochs L-BGFS-B per iteration                         & \multicolumn{2}{c|}{$5000$}            \\ \hline
L-BGFS-B maximum evaluations               & \multicolumn{2}{c|}{$20000 $}          \\ \hline
L-BGFS-B max corrections        & \multicolumn{2}{c|}{$50 $}             \\ \hline
L-BGFS-B max line search steps                  & \multicolumn{2}{c|}{$50 $  }           \\ \hline
L-BGFS-B tolerance                & $1e-6 $   & $1e-3$  \\ \hline
Number of training points $N_f$ & $20000 $  & $40400$        \\ \hline
Number of training points $N_u$  & $100$    & $20200$         \\ \hline
$\lambda_u$                             & \multicolumn{2}{c|}{$1$}               \\ \hline
$\lambda_f$                              & \multicolumn{2}{c|}{$10000 $  }        \\ \hline
%If mathematical scaling, mathematical scaling factor $\beta$                            & $1000 $              \\ \hline
%If physical scaling                             & \multicolumn{2}{c|}{Time unit changes from $h$ to $s$   }          \\ \hline
\end{tabular}
\label{tab:1D_hyper_param}
\end{table}

\subsection{Sensors' Optimization Models}

We use a mixed integer optimization model to find the optimal sensor placement inside power transformers to detect the temperature's stable points. A stable point is where the temperature changes the least over time. Therefore, it is defined where the absolute value of the time-averaged temperature change, given by the first-order partial derivatives with respect to space $\nabla u$, is at its minimum. For the 1D case, we consider $\frac{\partial u}{\partial x}$, while for the 2D case, we take the sum of the two partial derivatives with respect to $x$ and $y$, i.e., $\frac{\partial u}{\partial x} + \frac{\partial u}{\partial y}$.
The first-order partial derivatives are obtained when calculating the residual loss function MSE$_f$.
The goal is to place sensors at the stable points of the transformer. We set up a minimum and a maximum number of sensors, $n_{min}$ and $n_{max}$, respectively, that can be placed inside the transformers. 

In our study, we analyze three optimization models, which we will refer to throughout the paper as Model 1, Model 2, and Model 3 for simplicity. 
%Now assume that there are at most $n_{max}$ temperature sensors that can be placed inside the transformer, and there is a minimum requirement of $n_{min}$. To find the optimal placement of sensors to obtain the best temperature reconstruction under the limitation of the number of sensors, we use a mixed integer optimization model to find the temperature-stable points inside the transformer. A stable temperature point refers to the point where the absolute value of the time-averaged temperature change $\frac{\partial u}{\partial x}$ is minimum. This $\frac{\partial u}{\partial x}$ is obtained through $MSE_f$.

Model 1 is defined in the following way:

\begin{equation}
\label{eq:model1}
    \begin{aligned}
        \min_{\mathbf{s}} \quad & \mathbb{E}_{t \in D} \left| \nabla \cdot u( \mathbf{x}, t) \right| \cdot \mathbf{s}, \\
        \text{s.t.} \quad & \mathbf{s} = [s_1, s_2, \ldots, s_{N_x\cdot N_y}], \\
        & s_i \in \{0,1\}, \\
        & n_{\min} \leq \sum_i s_i \leq n_{\max},
    \end{aligned}
\end{equation}

\begin{comment}
\begin{equation}
     \label{model: basic}
    \begin{array}{cl}
        \displaystyle \min_{\mathbf{s}} & \displaystyle \mathbb{E}_{t \in D} \left| \nabla \cdot \left. u( \mathbf{x}, t) \right\vert_{\mathbf{x} = \bar{\mathbf{x}}} \right| \cdot \mathbf{s}, \\
        \text{s.t.} & \mathbf{s} = [s_1, s_2, \ldots, s_{N_x\cdot N_y}], \\
        & s_i \in \{0,1\}, \\
        &\displaystyle n_{\min} \leq \sum_i s_i \leq n_{\max},
    \end{array}
\end{equation}
\end{comment}

%\footnote{The divergence operator is defined as $\nabla \cdot f(\mathbf{x}) = \sum \frac{\partial f}{\partial x_i}$.}
where $\nabla \cdot$ is the divergence operator\footnote{The divergence operator is defined as $\nabla \cdot u(\mathbf{x}, t) = \frac{\partial u}{\partial x} + \frac{\partial u}{\partial y}$.}, $\mathbb{E}_{t \in D}$ refers to the mean operation over time with $D$ being a discrete set of time points. We also define a grid $\bar{\mathbf{x}}$ over $\bar{\Omega}_d$ as:
\[
    \bar{\Omega}_d = \{ \mathbf{x} \in \Omega \ | \ \text{distance to the boundary of } \Omega \text{ is more than } d \}
\]
with $N_x$ columns and $N_y$ rows. The binary variable $s_i$ indicates whether there is a sensor at the corresponding position $\mathbf{x}_i$. To be more clear,
\begin{equation*}
    s_i = \begin{cases} \begin{aligned}
        1, && \text{if there is a sensor at $\mathbf{x}_i$,} \\
        0, && \text{otherwise}.
    \end{aligned}\end{cases}
\end{equation*}

Model 1 is the basic optimization model, and it might cause the sensors to be clustered, not giving a good overall temperature representation inside the transformer. Therefore, an additional parameter is introduced to represent the minimum distance between two sensors, enforcing more sparsity. We define it as Model 2, and it is expressed as follows:
\begin{equation}
    \label{eq:model2}
    \begin{aligned}
        \min_{\mathbf{s}} \quad & \mathbb{E}_{t \in D} ( \left |\nabla \cdot u( \mathbf{x}, t) \right| )\cdot \mathbf{s} \\
        \text{s.t.} \quad  & \mathbf{s} = [s_1, s_2, \ldots, s_{N_x\cdot N_y}], \\
        & s_i \in \{0,1\}, \\
        & s_i + s_j \leq 1, \quad \text{if } \|\mathbf{x}_i-\mathbf{x}_j\| < d \quad \text{and} \quad \forall i, j, \ j \neq i, \\
        %& s_i = 0, \quad \text{if } \mathbf{x}_i \cdot \vec{x} < d, \quad \forall i \\
        & n_{\min} \leq \sum_i s_i \leq n_{\max},
    \end{aligned}
\end{equation}
%However, since the basic model may cause the sensors to be too clustered, an additional parameter is set to represent the minimum distance between two sensors to force the sensors to be sparser. So that the adjustment model is called Op.Model 1 and becomes
\begin{comment}
\begin{equation}
    \label{model: adj 1}
    \begin{array}{cl}
        \displaystyle \min_{\mathbf{s}} & % \displaystyle \frac{1}{N_t} \sum_{i=1}^{N_t} \left| \nabla_{\mathbf{x}} \cdot \left. T( \mathbf{x}, t_i) \right\vert_{\mathbf{x} = \bar{\mathbf{x}}} \right| \cdot \mathbf{s} \\
        \mathbb{E}_{t \in D} ( \left |\nabla_{\mathbf{x}} \cdot \left. u( \mathbf{x}, t) \right\vert_{\mathbf{x} = \bar{\mathbf{x}}} \right| )\cdot \mathbf{s} \\
        \text{s.t.} & \mathbf{s} = [s_1, s_2, \ldots, s_{N_x\cdot N_y}], \\
        & s_i \in \{0,1\}, \\
        & s_i + s_j \leq 1, \quad \text{if } \|\mathbf{x}_i-\mathbf{x}_j\| < d \quad \text{and} \quad \forall i, j, \ j \neq i, \\
        %& s_i = 0, \quad \text{if } \mathbf{x}_i \cdot \vec{x} < d, \quad \forall i \\
        &\displaystyle n_{\min} \leq \sum_i s_i \leq n_{\max},
    \end{array}
\end{equation}
\end{comment}
The setting of $d$ depends on the user's "experience". Since sensors cannot be placed at a distance less than $d$, this forced placement may cause the sensors to miss important information if $d$ is set too large. To ensure that the sensors are placed in a position that can monitor temperature at key locations inside the transformer and collect sufficient information simultaneously, an additional distance parameter $d_1$ is included in the optimization model to ensure that the sensors are spread out to a certain extent. Therefore, the distance $d$ describes the distance that must be maintained between the sensors, which also depends on the sensor size. The distance $d_1$ is another limit between two sensors, which describes the measurement overlap caused by the distance between the two sensors, which can cause information waves. The corresponding waste of the two sensors is defined as the cost $c_i^j$, that is,
\begin{equation}
    c_i^j = \begin{cases}
        \mathbb{E}_{t \in D} \left(\nabla \cdot u (\mathbf{x}, t) \right) \left( d_1 - \|\mathbf{x}_i - \mathbf{x}_j\| \right) & \text{if } j \neq i, \\
        0 & \text{if } i = j.
    \end{cases}
\end{equation}

\begin{comment}
\begin{equation}
    c_i^j = \begin{cases}
        \mathbb{E}_{t \in D} \left( \left. \nabla_{\mathbf{x}} \cdot u (\mathbf{x}, t) \right|_{\mathbf{x} = \mathbf{x}_i} \right) \left( d_1 - \|\mathbf{x}_j - \mathbf{x}\| \right) & \text{if } j \neq i, \\
        0 & \text{if } i = j.
    \end{cases}
\end{equation}
\end{comment}

With the help of the big-M formulation, the final optimization model, which we define as Model 3, becomes:
\begin{equation}
    \begin{aligned}
        \min_{\mathbf{s}} \quad & \mathbb{E}_t ( \left |\nabla_{\mathbf{x}} \cdot  u( \mathbf{x}, t) \right| )\cdot \mathbf{s} + \sum_i \mathcal{L}_i \\
        \text{s.t. } \quad & \mathbf{s} = [s_1, s_2, \cdots, s_{N_x \cdot N_y}],  \\
        & s_i +s_j \leq 1, \quad \text{if } \| \mathbf{x}_i - \mathbf{x}_j \|<d \quad \& \quad \forall i, j, \ j \neq i, \\
        & n_{min} \leq \sum_i s_i \leq n_{max}, \\
        & s_i \in \{0,1\}, \quad \text{for } i = 1, \cdots, N_x \cdot N_y, \\
        & c_i = \sum_{j=1}^{N_x} s_j c_i^j, \quad \text{for } i = 1, \cdots, N_x \cdot N_y, \\
         & c_i - M(1 - s_i) \leq \mathcal{L}_i \leq c_i + M(1 - s_i),  \quad \text{for } i = 1, \cdots, N_x\cdot N_y, \\
         & \mathcal{L}_i \leq M s_i,  \quad \text{for } i = 1, \cdots, N_x \cdot N_y, \\
    \end{aligned}
\end{equation}
\begin{comment}
\begin{equation}
    \begin{array}[t]{cl}
        \displaystyle \min_{\mathbf{s}} & % \displaystyle \frac{1}{N_t} \sum_{i=1}^{N_t} \left| \left. \nabla_{\mathbf{x}} \cdot T(\mathbf{x}, t_i) \right|_{\mathbf{x} = \bar{\mathbf{x}}} \right| \cdot \mathbf{s} + \sum_i \mathcal{L}_i \\
        \mathbb{E}_t ( \left |\nabla_{\mathbf{x}} \cdot \left. u( \mathbf{x}, t) \right\vert_{\mathbf{x} = \bar{\mathbf{x}}} \right| )\cdot \mathbf{s} + \sum_i \mathcal{L}_i \\
        \text{s.t. } & \mathbf{s} = [s_1, s_2, \cdots, s_{N_x \cdot N_y}],  \\
        & s_i +s_j \leq 1, \quad \text{if } \| \mathbf{x}_i - \mathbf{x}_j \|<d \quad \& \quad \forall i, j, \ j \neq i, \\
        & n_{min} \leq \sum_i s_i \leq n_{max}, \\
        & s_i \in \{0,1\}, \quad \text{for } i = 1, \cdots, N_x \cdot N_y, \\
        & c_i = \sum_{j=1}^{N_x} s_j c_i^j, \quad \text{for } i = 1, \cdots, N_x \cdot N_y, \\
         & c_i - M(1 - s_i) \leq \mathcal{L}_i \leq c_i + M(1 - s_i),  \quad \text{for } i = 1, \cdots, N_x\cdot N_y, \\
         & \mathcal{L}_i \leq M s_i,  \quad \text{for } i = 1, \cdots, N_x \cdot N_y, \\
    \end{array}
\end{equation}
\end{comment}
where $\mathcal{L}_i$ represents the cost of placing a sensor at position $\mathbf{x}_i$, and the penalty coefficient $M$ is selected as $1000$.

\section{Results}

This section reports the results obtained to model the temperature inside a power transformer in 1D and 2D and the corresponding optimal sensor placement model.
The 1D model is run for 20000 epochs, 10000 using Adam optimizer, and 10000 with L-BFGS-B, taking approximately 45 minutes using GPUs from Google Colab \cite{bisong2019google}.
%The 1D model is run for 11 full training iterations, each taking approximately 45 minutes. We compare the results obtained after the first and last training iterations along the entire section for 1D and 2D problems to better understand the PINN model's performance and convergence. 
The hyperparameters used for the model are listed in Table \ref{tab:1D_hyper_param}.

\begin{figure}[ht]
     \centering
     \begin{subfigure}[b]{1\textwidth}
         \centering
         \includegraphics[width=0.8\textwidth]{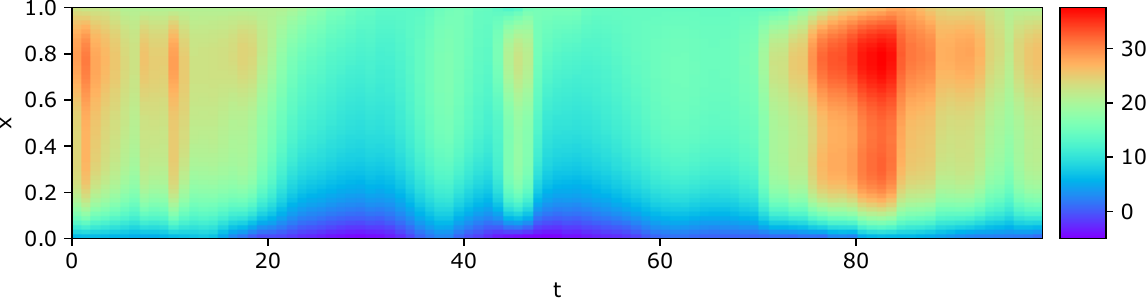}
         \caption{Comsol solution.}
         \label{fig:1d_comsol}
     \end{subfigure}
     %\hfill
     \begin{subfigure}[b]{1\textwidth}
         \centering
         \includegraphics[width=0.8\textwidth]{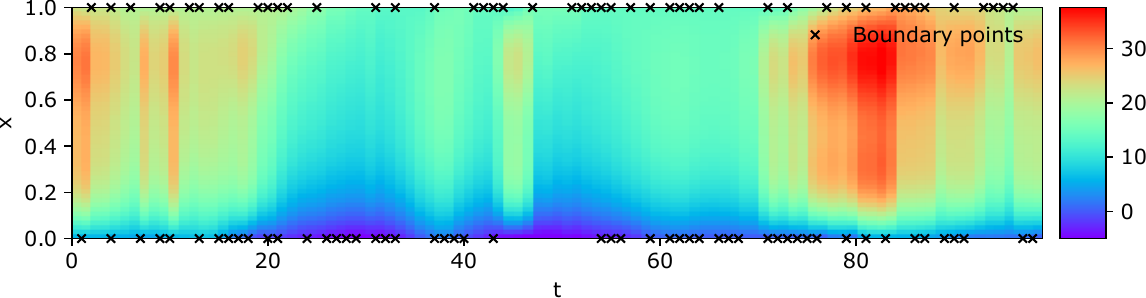}
         \caption{PINN solution.}
         \label{fig:1d_pinn}
     \end{subfigure}
\caption{Solution of the first 100 hours for the 1D problem using Comsol and PINN.}
\label{fig:1d_solutions_all}
\end{figure}

Figure \ref{fig:1d_solutions_all} shows the solution for the 1D problem for the first 100 hours of the dataset. In particular, Figure \ref{fig:1d_comsol} represents the reference solution calculated using Comsol, while Figure \ref{fig:1d_pinn} shows the results obtained with PINNs. 

\begin{figure}[ht]
    \centering
    \includegraphics[width=\linewidth]{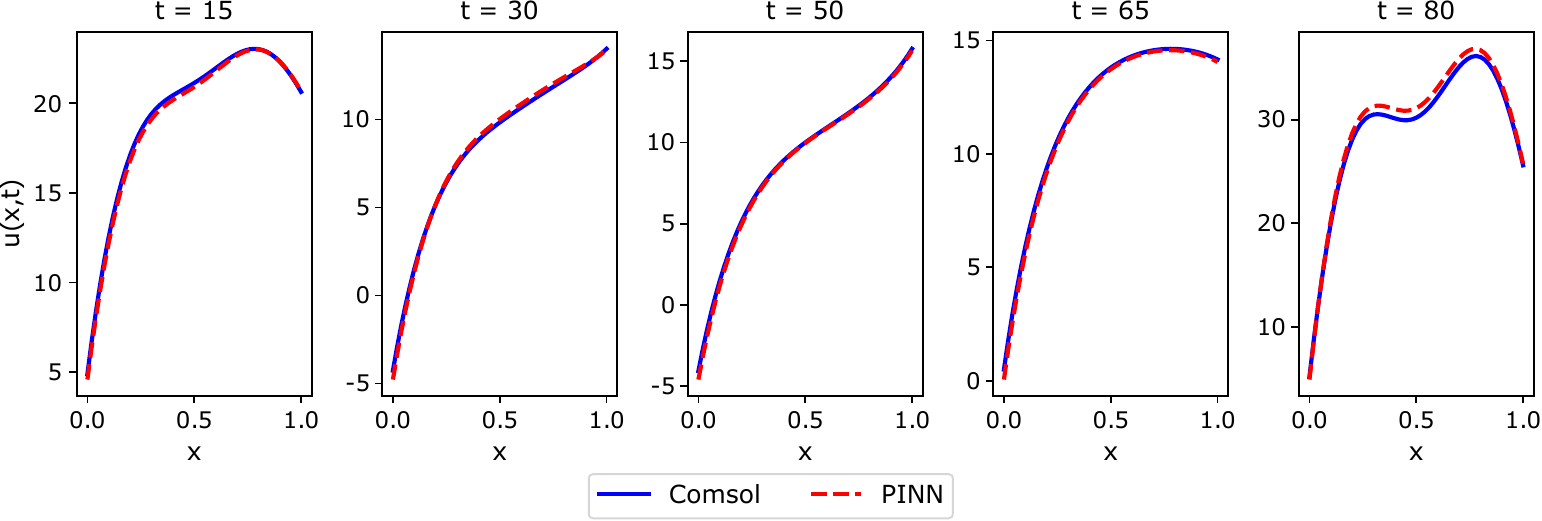}
    \caption{Comparison of the solution obtained by Comsol (blue line) and PINN (red-dotted line) for the 1D problem at several time points.}
    \label{fig:comparison_1d}
\end{figure}

\begin{comment}
\begin{figure}[ht]
     \centering
     \begin{subfigure}[b]{1\textwidth}
         \centering
         \includegraphics[width=0.8\textwidth]{figure_pdf/Temp-figure-SameScale/1D-PINN/1D_sin_comsol.pdf}
         \caption{Comsol solution.}
         \label{fig:comsol_1d}
     \end{subfigure}
     %\hfill
     \begin{subfigure}[b]{1\textwidth}
         \centering
         \includegraphics[width=0.8\textwidth]{figure_pdf/Temp-figure-SameScale/1D-PINN/1D_sin_1train_PINN.pdf}
         \caption{PINN solution from the first training iteration.}
         \label{fig:pinn_first_iter}
     \end{subfigure}
     %\hfill
     \begin{subfigure}[b]{1\textwidth}
         \centering
         \includegraphics[width=0.8\textwidth]{figure_pdf/Temp-figure-SameScale/1D-PINN/1D_sin_11train_PINN.pdf}
         \caption{PINN solution from the last training iteration.}
         \label{fig:pinn_last_iter}
     \end{subfigure}
\caption{Solution of the first 100 hours for the 1D problem using Comsol and PINN after the first and the last training iterations.}
\label{fig: 1D result}
\end{figure}
\end{comment}
We can notice that the PINN solution is already very close to the reference one. We can compare the results more clearly by looking at the plots in Figure \ref{fig:comparison_1d}, where we consider five specific times, i.e., $t=15$, $t=30$, $t=50$, $t=65$, and $t=80$. The blue lines represent the reference solution using Comsol, and the red-dotted lines represent the PINN solution. PINNs capture almost perfectly the solution, especially for the first time steps. For $t=80$, there is a slight shift from the reference solution for the PINN model with a minimal difference. 

\begin{figure}
    \centering
    \includegraphics[width=0.6\linewidth]{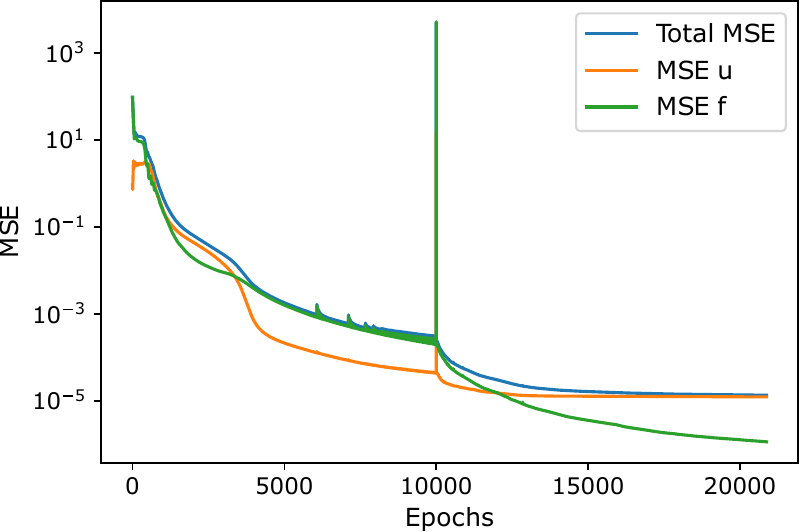}
    \caption{Loss functions for the 1D PINN model.}
    \label{fig:loss_functions_1D}
\end{figure}

\begin{figure}[ht]
     \centering
     \begin{subfigure}[b]{0.45\textwidth}
         \centering
         \includegraphics[width=\textwidth]{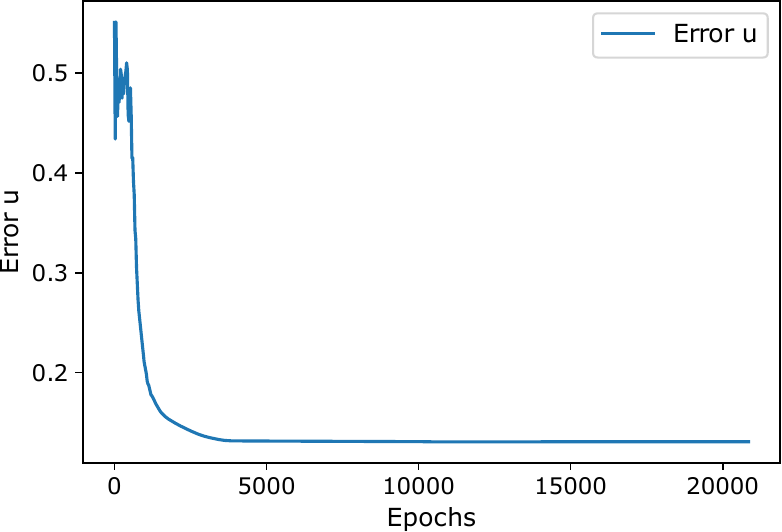}
         \caption{Overall temperature distribution.}
         \label{fig:error_u_1D}
     \end{subfigure}
     \hfill
     \begin{subfigure}[b]{0.45\textwidth}
         \centering
         \includegraphics[width=\textwidth]{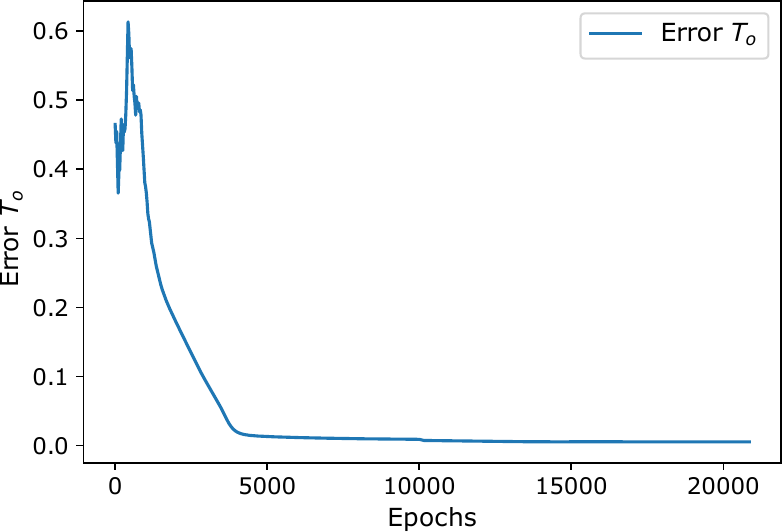}
         \caption{Top-oil temperature $T_o$}
         \label{fig:error_Ttop_1D}
     \end{subfigure}
\caption{Relative $L_2$ errors for the $1D$ problem between the reference solution given by Comsol and the PINN solution.}
\label{fig:errors_all_1d}
\end{figure}

Figure \ref{fig:loss_functions_1D} shows the evolution of the loss function over the epochs.
The evolution of the relative $L_2$ errors is shown in Figure \ref{fig:errors_all_1d}. Figure \ref{fig:error_u_1D} represents the relative $L_2$ error between the model’s temperature prediction and the reference after each epoch. Figure \ref{fig:error_Ttop_1D} shows the same but only for the top-oil temperature $T_o$. For both cases, the errors decrease smoothly, reaching a value of $1.306\cdot10^{-1}$ for the overall temperature distribution and $5.532\cdot10^{-3}$ for $T_o$. 
%Figures \ref{fig:error_u_1D} and \ref{fig:err} show the $L_2$ error between the model’s temperature prediction and the reference after each epoch in the first and the last training iteration. It can be seen that although the lowest error value in the first training is approximately $2\cdot10^{-1}$ in the middle of the training, the error increases again in the following epochs, converging to about $4\cdot10^{-1}$. The $L_2$ error for the last training iteration instead gets lower and decreases for the whole training, reaching a value of about $1.22\cdot10^{-1}$.

\begin{figure}[ht]
     \centering
     \begin{subfigure}[b]{0.49\textwidth}
         \centering
         \includegraphics[width=\textwidth]{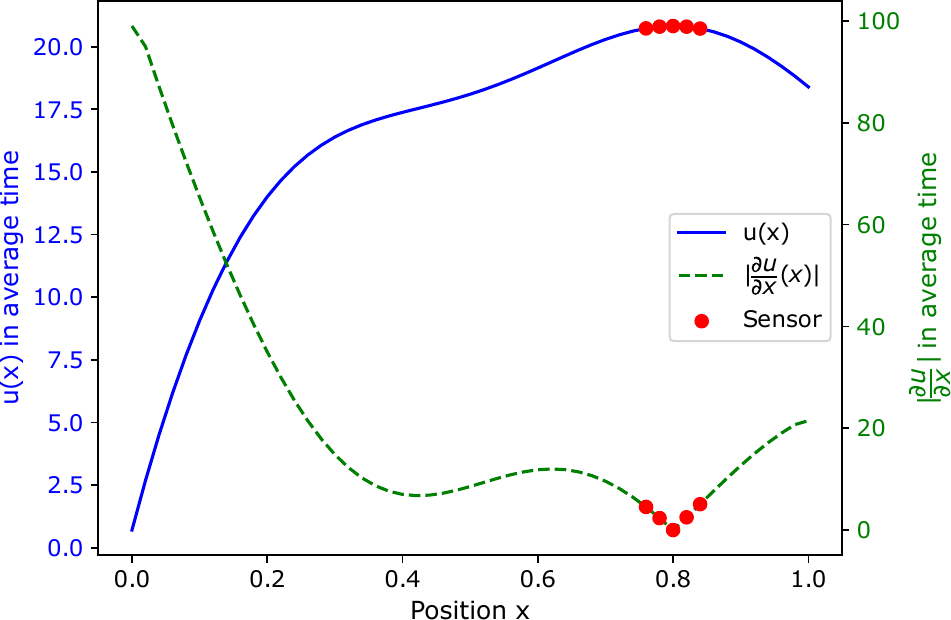}
         \caption{Model 1}
         \label{fig:1d_basicmodel}
     \end{subfigure}
     \begin{subfigure}[b]{0.49\textwidth}
         \centering
         \includegraphics[width=\textwidth]{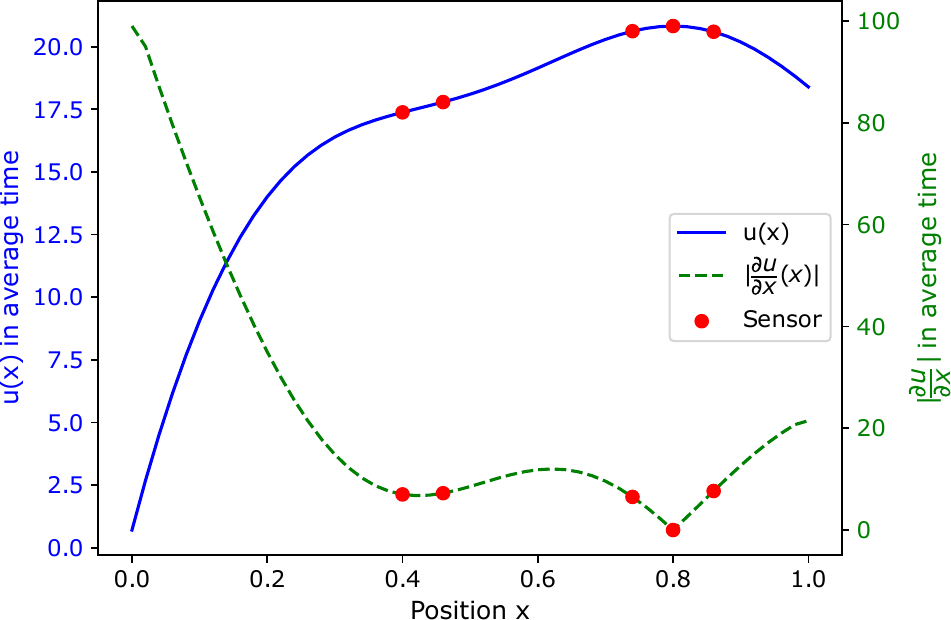}
         \caption{Model 2 with $d=0.05$}
         \label{fig:1d_model_distance}
    \end{subfigure}
    \hfill
     \begin{subfigure}[b]{0.49\textwidth}
         \centering
         \includegraphics[width=\textwidth]{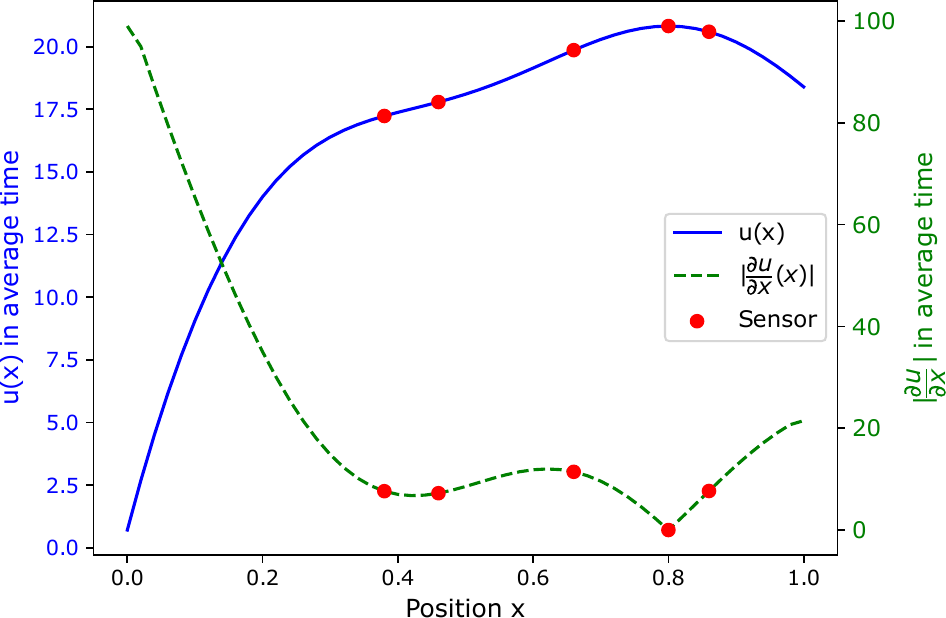}
         \caption{Model 3 with $d=0.05, d_1=0.2$}
         \label{fig:1d_model_distance_penalties}
     \end{subfigure}
\caption{The optimal sensor placement for 1D problem, with $n_{min}=5$ and $n_{max}=10$.}
\label{fig:op_models_1d}
\end{figure}

We now analyze the results obtained by the optimization model to find the optimal positions for multiple sensors. We set the minimum and maximum number of sensors as $n_{min}=5$ and $n_{max}=10$, respectively.
The results are shown in Figure \ref{fig:op_models_1d}. Each plot shows the time-averaged temperature results (the blue lines) and the time-averaged first-order spatial derivatives of the temperature (the green-dotted lines) for a more explicit representation of the sensor placement. 
In particular, Figure \ref{fig:1d_basicmodel} shows the sensor placement using Model 1, and Figures \ref{fig:1d_model_distance} and \ref{fig:1d_model_distance_penalties} represent the results for Model 2 and Model 3, respectively. For Model 2, the distance used is $d=0.05$; similarly, Model 3 uses $d=0.05$ with the additional distance parameter value $d_1=0.2$. The sensors are mainly located around the stable points, i.e., the positions where the temperature has the least change over time. In Model 1, the sensors are all placed close to each other, meaning that their information overlaps as they do not cover enough space range. This problem is overcome by the other two optimization models, where the sensors are more spread out within the region of interest with the help of the distance parameters and the inclusion of the penalty parameter for Model 3. The sensor placement of the two optimized models is still concentrated around the stable points; however, Model 3 has a better distribution of the sensors, keeping more distance between them, which makes it more practical to use when detecting the temperature distribution inside a power transformer. 

Moving to the 2D problem, the complexity increases, making the training exponentially more expensive compared to the 1D counterpart. Moreover, the training time also increases. With the same hyperparameters, training the 2D problem takes approximately 60 minutes. The hyperparameters utilized are reported in Table \ref{tab:1D_hyper_param}. 

\begin{figure}[ht]
     \centering
     \begin{subfigure}[b]{0.4\textwidth}
         \centering
         \includegraphics[width=\textwidth]{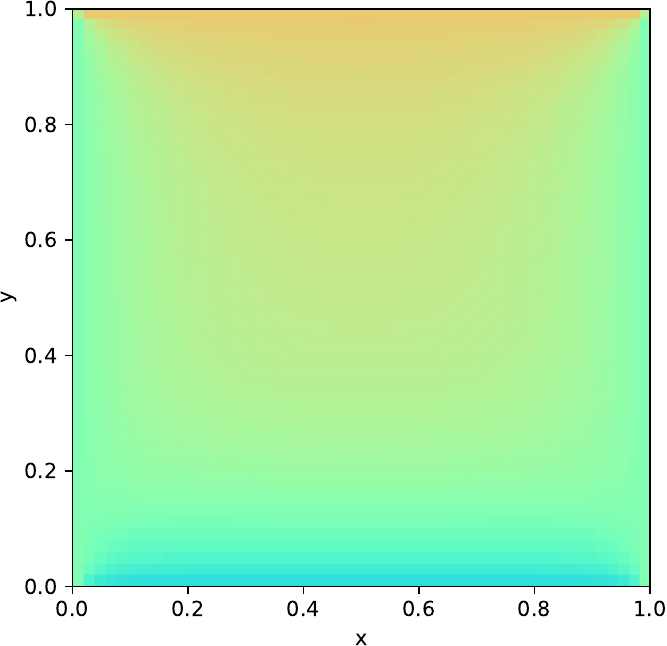}
         \caption{Comsol, $t=10$.}
         \label{fig:2d_comsol_t10}
     \end{subfigure}
     %\hfill
     \begin{subfigure}[b]{0.4\textwidth}
         \centering
         \includegraphics[width=\textwidth]{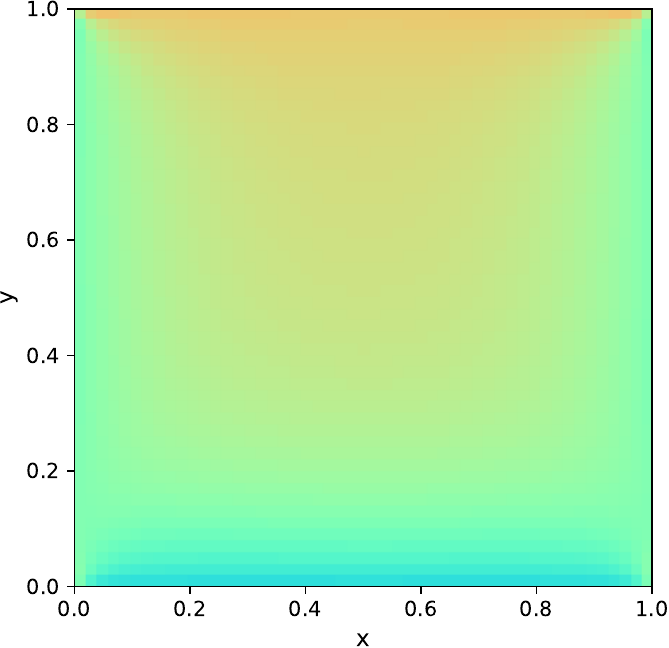}
         \caption{PINN, $t=10$.}
         \label{fig:2d_pinn_t10}
     \end{subfigure}
     %\hfill
     \begin{subfigure}[b]{0.4\textwidth}
         \centering
         \includegraphics[width=\textwidth]{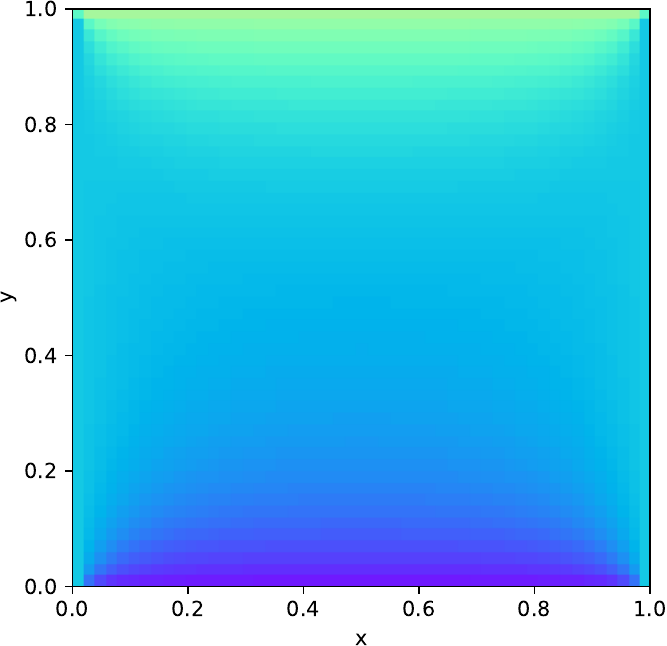}
         \caption{Comsol, $t=50$.}
         \label{fig:2d_comsol_t50}
     \end{subfigure}
          %\hfill
     \begin{subfigure}[b]{0.4\textwidth}
         \centering
         \includegraphics[width=\textwidth]{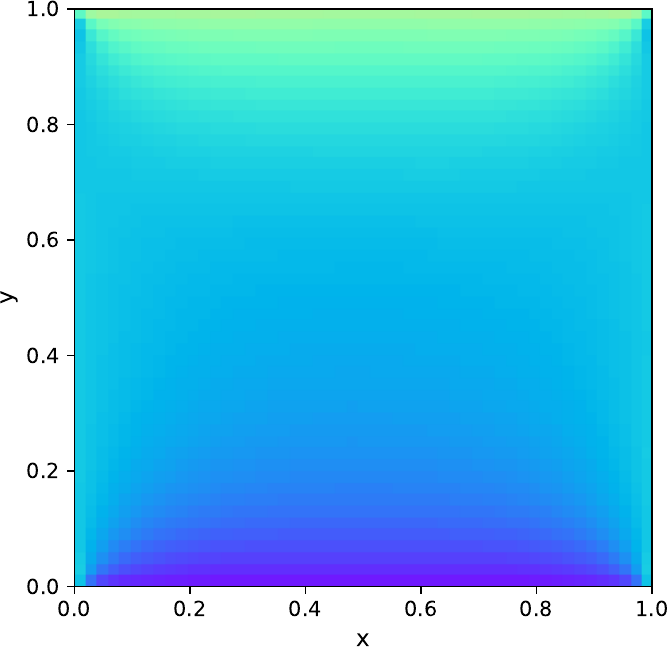}
         \caption{PINN, $t=50$.}
         \label{fig:2d_pinn_t50}
     \end{subfigure}
          %\hfill
     \begin{subfigure}[b]{0.4\textwidth}
         \centering
         \includegraphics[width=\textwidth]{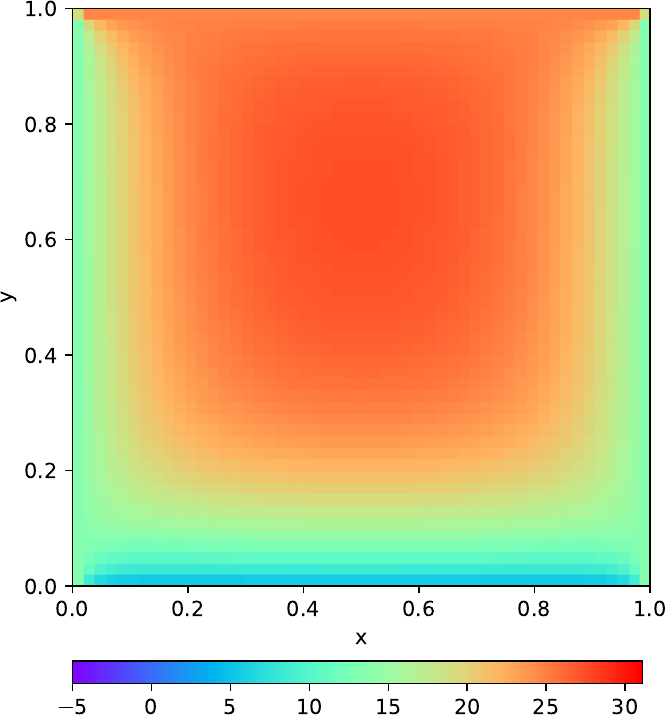}
         \caption{Comsol, $t=80$.}
         \label{fig:2d_comsol_t80}
     \end{subfigure}
          %\hfill
     \begin{subfigure}[b]{0.4\textwidth}
         \centering
         \includegraphics[width=\textwidth]{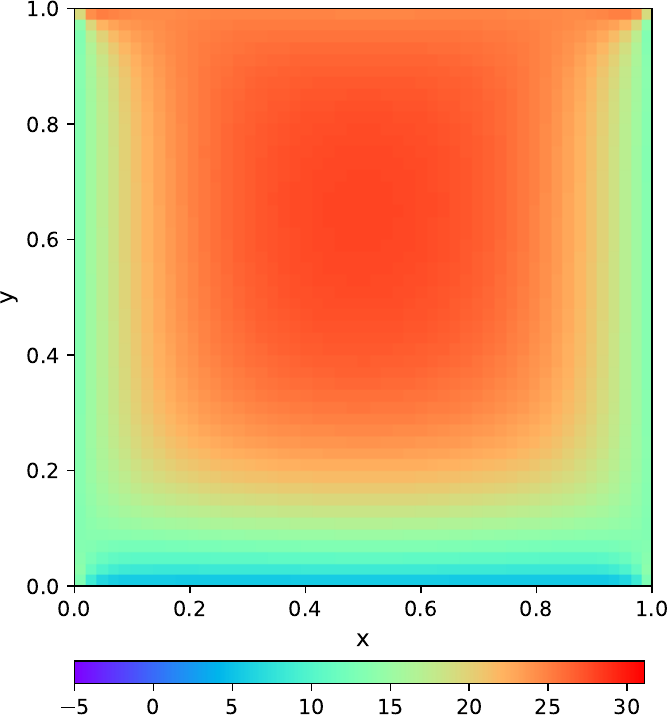}
         \caption{PINN, $t=80$.}
         \label{fig:2d_pinn_t80}
     \end{subfigure}
\caption{Solution for the 2D problem using Comsol (left) and PINNs (right) for $t=10$, $t=50$, and $t=80$.}
\label{fig:2d_solution_all}
\end{figure}

\begin{figure}[ht]
     \centering
     \begin{subfigure}[b]{0.9\textwidth}
         \centering
         \includegraphics[width=\textwidth]{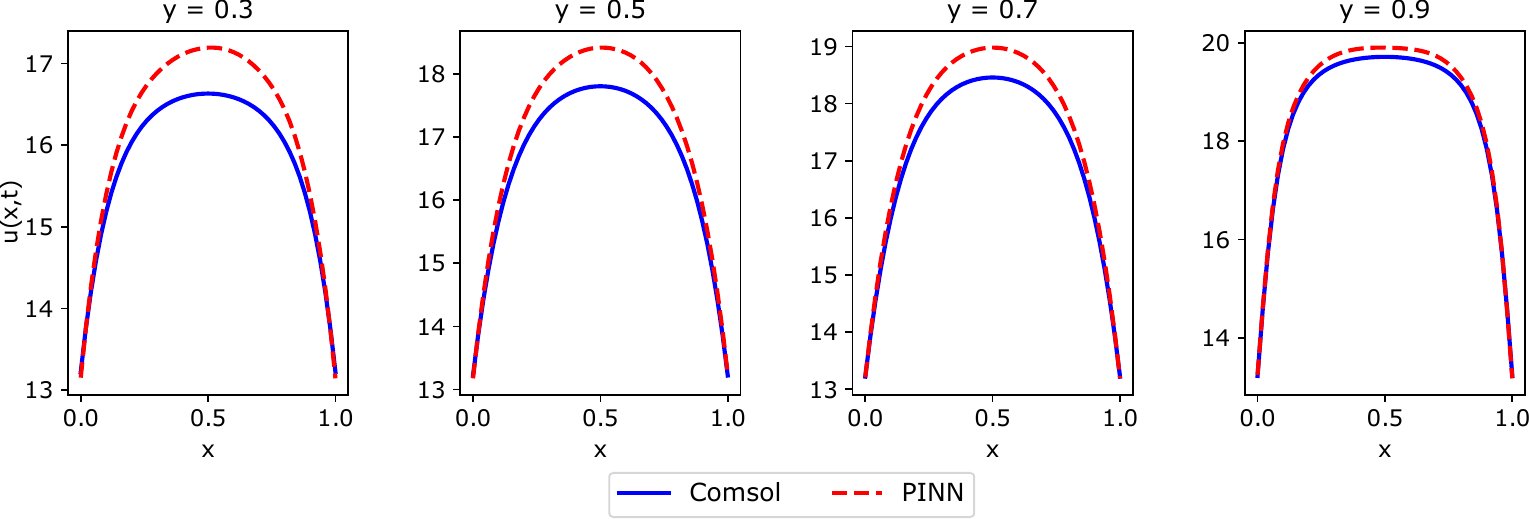}
         \caption{$t=10$.}
         \label{fig:2d_comparison_t10}
     \end{subfigure}
     \hfill
     \begin{subfigure}[b]{0.9\textwidth}
         \centering
         \includegraphics[width=\textwidth]{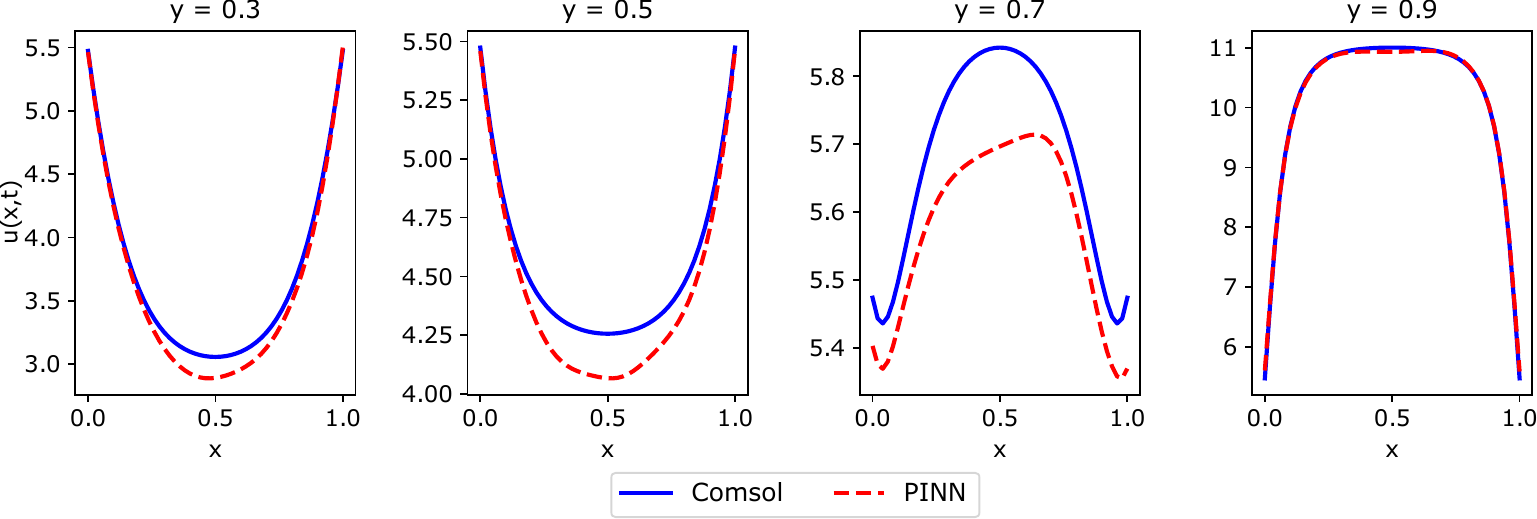}
         \caption{$t=50$.}
         \label{fig:2d_comparison_t50}
     \end{subfigure}
          \hfill
     \begin{subfigure}[b]{0.9\textwidth}
         \centering
         \includegraphics[width=\textwidth]{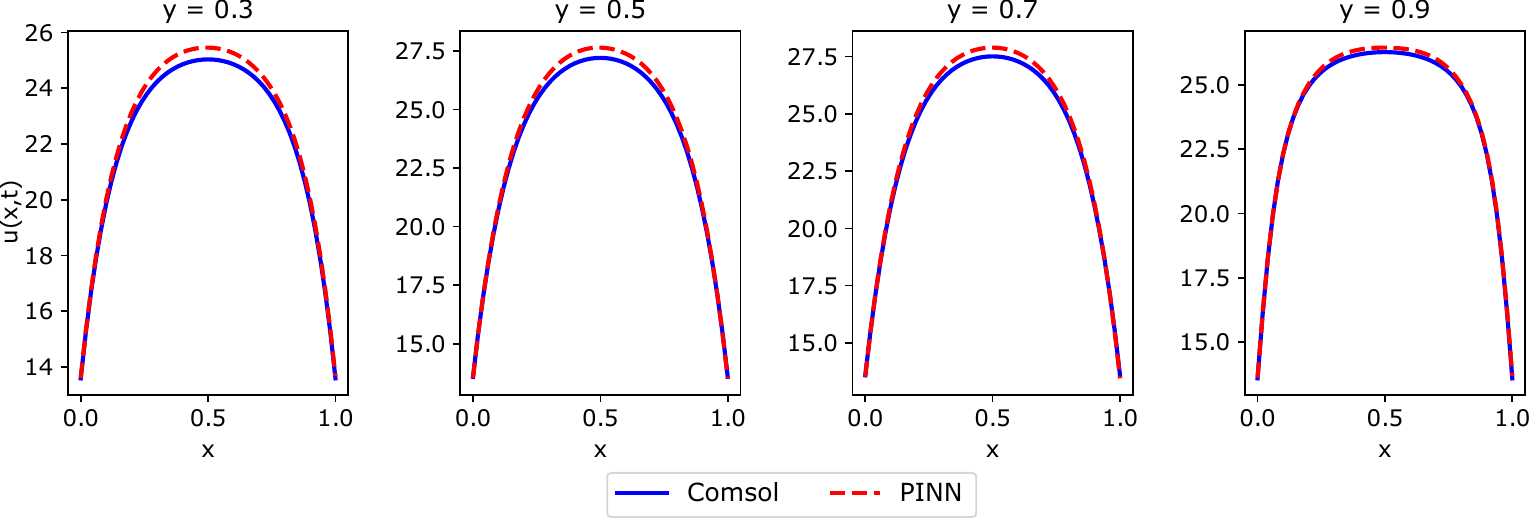}
         \caption{$t=80$.}
         \label{fig:2d_comparison_t80}
     \end{subfigure}
\caption{Comparison of the solution obtained by Comsol (blue line) and PINNs (red-dotted line) for the 2D problem at specific $y$ location and $t=10$, $t=50$, and $t=80$.}
\label{fig:2d_solution_compare_all}
\end{figure}

To show the results for the 2D problem, we pick three arbitrary time points: $t=10$, $t=50$, and $t=80$. Figure \ref{fig:2d_solution_all} shows the plots for these time points. In particular, Figures \ref{fig:2d_comsol_t10}, \ref{fig:2d_comsol_t50} and \ref{fig:2d_comsol_t80} show the reference solution obtained with Comsol, while Figures \ref{fig:2d_pinn_t10}, \ref{fig:2d_pinn_t50} and \ref{fig:2d_pinn_t80} display the PINN solutions. We can notice that the model predicts the temperature distribution quite accurately for all three time steps compared to the Comsol solution. In Figure \ref{fig:2d_solution_compare_all} we can look at a closer comparison between the reference solution, the blue lines, and the PINN solution, the red-dotted lines. The comparisons are for the three time points, and we took four random locations for the $y$ values, i.e., $y=0.3$, $y=0.5$, $y=0.7$, and $y=0.9$. In particular, Figures \ref{fig:2d_comparison_t10}, \ref{fig:2d_comparison_t50}, and \ref{fig:2d_comparison_t80} show the results for $t=10$, $t=50$, and $t=80$, respectively. We can notice slight discrepancies between the solutions, especially for the first three $y$ locations for $t=10$ and $t=50$. Overall, the solutions are already notable, given the amount of training and the number of training and collocation points used. However, to obtain more accurate and reliable estimations with PINNs, it is necessary to train longer and use more training points. 

\begin{figure}[ht]
    \centering
    \includegraphics[width=0.6\linewidth]{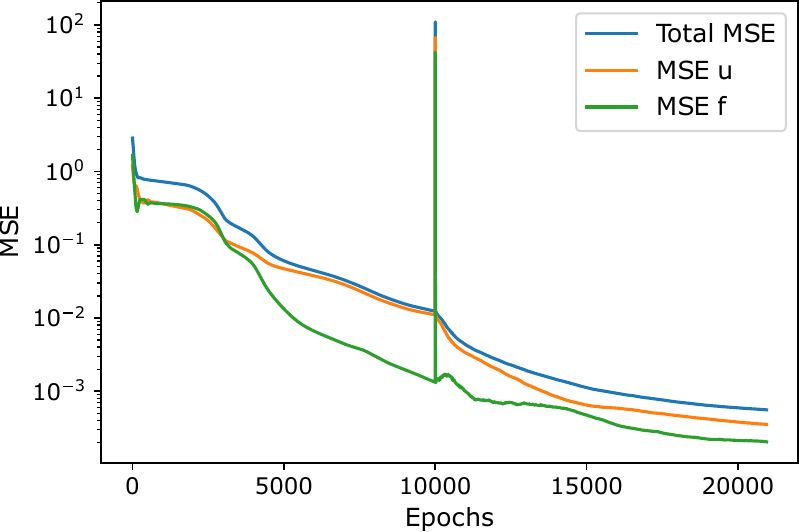}
    \caption{Loss functions for the 2D PINN model.}
    \label{fig:loss_functions_2D}
\end{figure}

\begin{figure}[ht]
     \centering
     \begin{subfigure}[b]{0.45\textwidth}
         \centering
         \includegraphics[width=\textwidth]{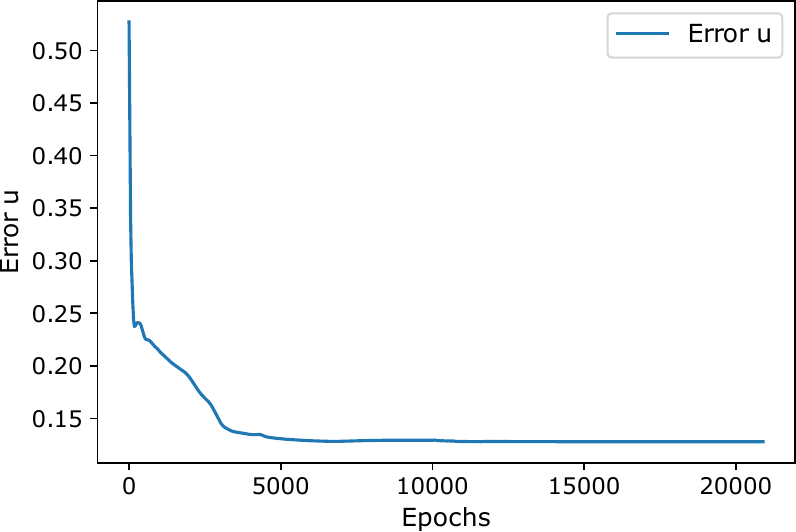}
         \caption{$L_2$ error of the overall temperature solution.}
         \label{fig:error_u_2D}
     \end{subfigure}
     \hfill
     \begin{subfigure}[b]{0.45\textwidth}
         \centering
         \includegraphics[width=\textwidth]{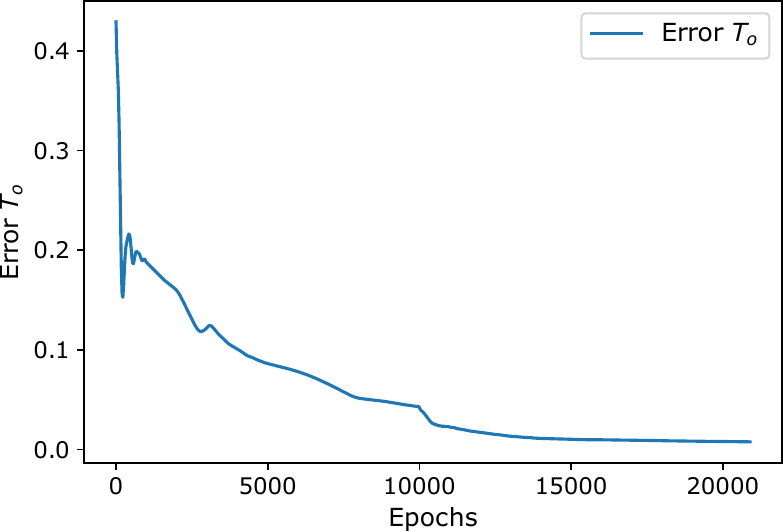}
         \caption{$L_2$ error of the $T_o$ solution.}
         \label{fig:error_Ttop_2D}
     \end{subfigure}
\caption{$L_2$ error for the $2D$ problem between the reference solution given by Comsol and the PINN solution.}
\label{fig:errors_all_2d}
\end{figure}

\begin{comment}
\begin{figure}[ht]
    \centering
    \includegraphics[width=0.6\linewidth]{New_figure/PINN/2D_crop/2D_all_MSE-crop.pdf}
    \caption{Loss functions for the 2D PINN model.}
    \label{fig:loss_functions_2D}
\end{figure}
\begin{figure}[ht]
     \centering
     \begin{subfigure}[b]{0.45\textwidth}
         \centering
         \includegraphics[width=\textwidth]{New_figure/PINN/2D_crop/2D_Error_temp-crop.pdf}
         \caption{$L_2$ error of the overall temperature solution.}
         \label{fig:error_u_2D}
     \end{subfigure}
     \hfill
     \begin{subfigure}[b]{0.45\textwidth}
         \centering
         \includegraphics[width=\textwidth]{New_figure/PINN/2D_crop/2D_Error_u_top-crop.pdf}
         \caption{$L_2$ error of the $T_o$ solution.}
         \label{fig:error_Ttop_2D}
     \end{subfigure}
\caption{$L_2$ error for the $2D$ problem between the reference solution given by Comsol and the PINN solution.}
\label{fig:errors_all_2d}
\end{figure}
\end{comment}
We can look at Figure \ref{fig:loss_functions_2D}, where the loss functions are plotted to see that the training might not be enough at this stage. Similarly to the 1D case, we use 10000 epochs with Adam optimizer and 10000 epochs with L-BFGS-B. From the plot, we can notice that the training with Adam has still not properly converged after 10000 epochs, therefore, longer training is required to converge to the optimal solution. Figure \ref{fig:errors_all_2d} shows the evolution of the relative $L_2$ errors for the overall solution $u$ and the top-oil temperature $T_o$. Figure \ref{fig:error_u_2D} shows the error after each epoch for the temperature distribution over the whole domain, which reaches a value of $1.278\cdot10^{-1}$ at the end of the training. Figure \ref{fig:error_Ttop_2D} represents the error for the top-oil temperature, achieving a value of $7.577\cdot10^{-3}$. 

\begin{figure}[ht]
     \centering
     \begin{subfigure}[b]{\textwidth}
         \centering
         \includegraphics[width=0.8\textwidth]{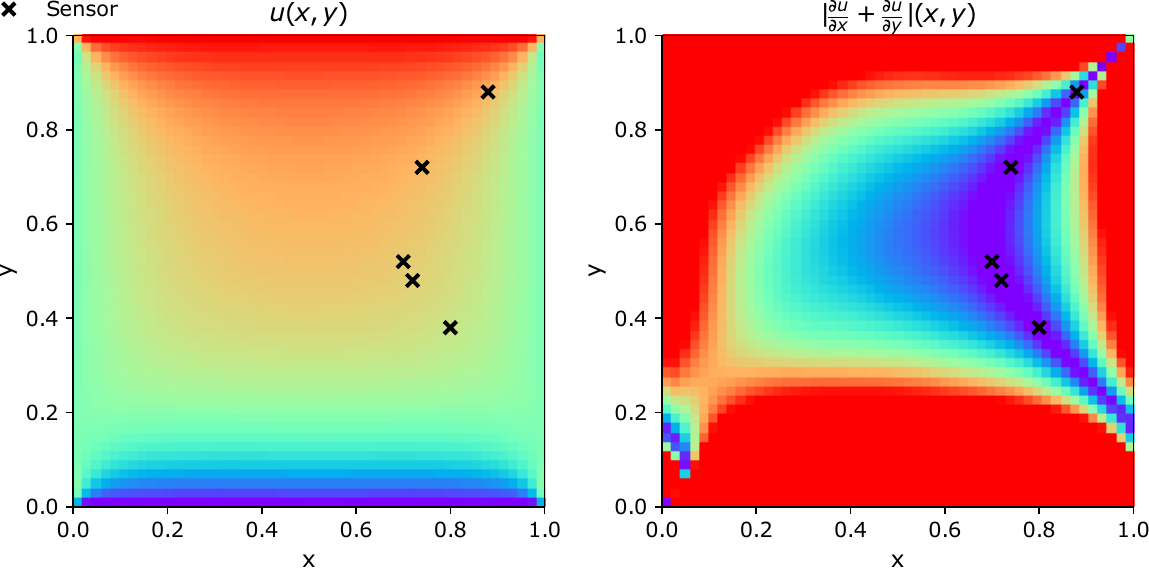}
         \caption{Model 1}
         \label{fig:2D_basic_opt}
     \end{subfigure}
     \begin{subfigure}[b]{\textwidth}
         \centering
         \includegraphics[width=0.8\textwidth]{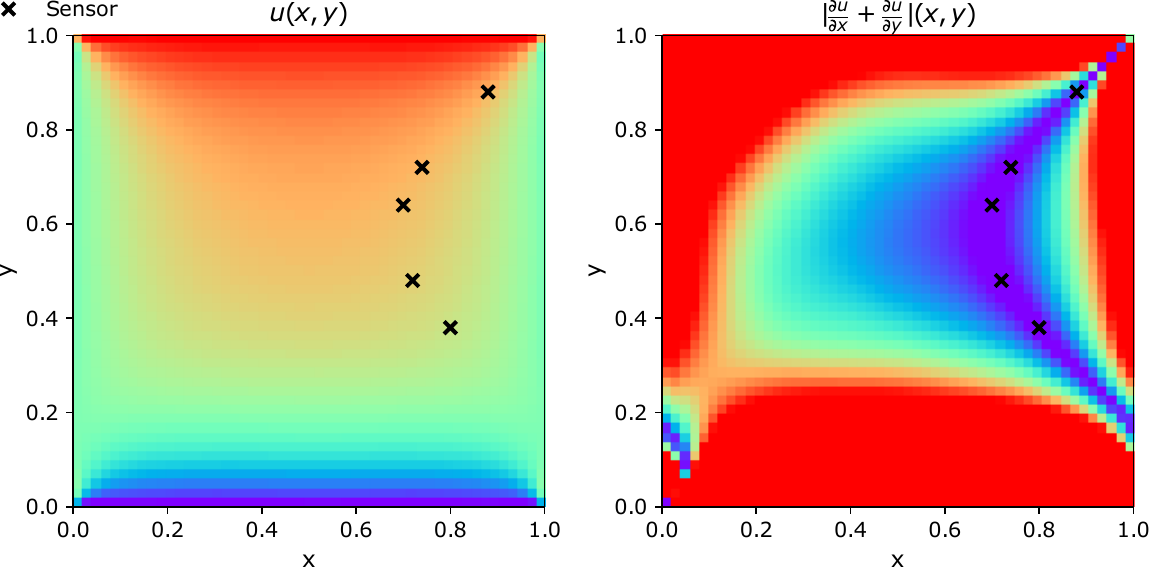}
         \caption{Model 2 with $d=0.05$}
         \label{fig:2D_model1_opt}
    \end{subfigure}
    \hfill
     \begin{subfigure}[b]{\textwidth}
         \centering
         \includegraphics[width=0.8\textwidth]{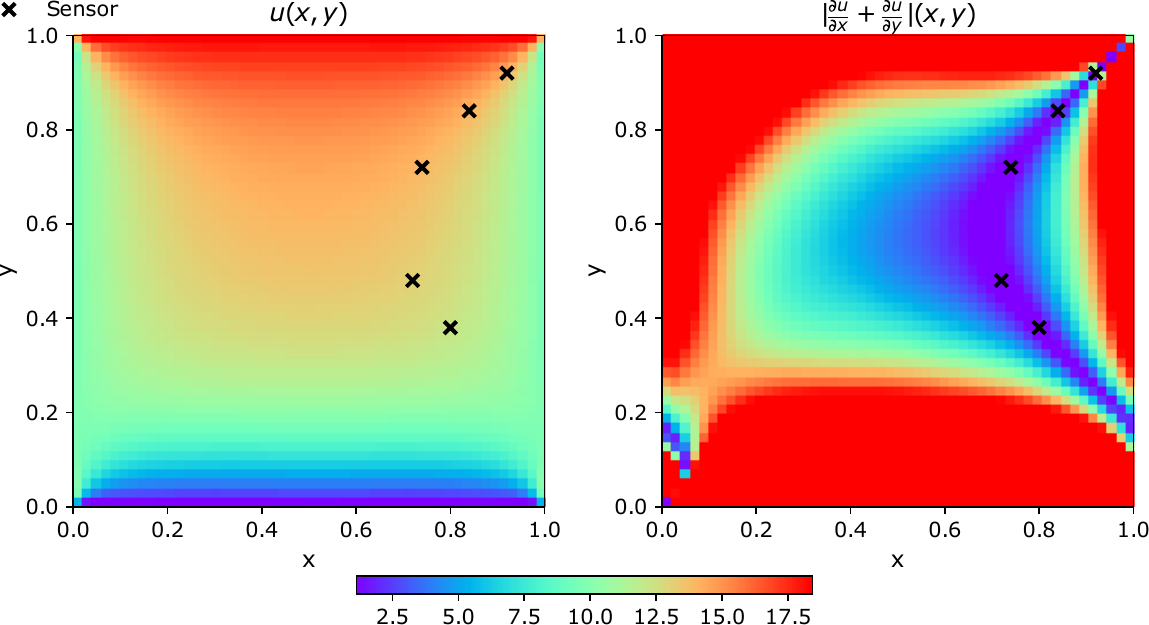}
         \caption{Model 3 with $d=0.05, d_1=0.2$}
         \label{fig:2D_model2_opt}
     \end{subfigure}
\caption{The optimal sensor placement for the 2D problem, with $n_{min} = 5$, $n_{max}=10$.}
\label{fig:2D_sensor_op_all}
\end{figure}

Given the larger domain, the complexity of finding the optimal placement of sensors in the 2D problem increases compared to the 1D one. Figure \ref{fig:2D_sensor_op_all} shows the results obtained setting up the minimum and the maximum number of sensors as $n_{min}=5$ and $n_{max}=10$. On the left plots, the time-averaged temperature results are displayed, and, in addition, on the right plots, the sum of the time-averaged first-order spatial derivatives of the temperature $\Big|\frac{\partial u}{\partial x} + \frac{\partial u}{\partial y}\Big|(x,y)$ is also shown to represent the results more precisely. We investigate similar cases for the optimized models as for the 1D problem. Figure \ref{fig:2D_basic_opt} shows the sensor placement with Model 1, while Figures \ref{fig:2D_model1_opt} and \ref{fig:2D_model2_opt} display Model 2 and Model 3 results, respectively. As for the 1D problem, we take $d=0.05$ for both optimized models, adding the distance parameter value $d_2=0.2$ for Model 3. 
%It can be seen that the sensors are placed near $\mathbb{E}_t ( \nabla_{\mathbf{x}} \cdot T( \mathbf{x}, t_i) \vert_{\mathbf{x} = \bar{\mathbf{x}}}  )=0$. 
As we introduce penalties for distance and temperature gradient, the dispersion of the sensors can be controlled more while ensuring that the significant temperatures, i.e., the stable points, are considered. 

\section{Discussion and Conclusions}

Physics-Informed Neural Networks show many benefits for predicting internal temperatures of power components such as power transformers. However, with higher spatial dimensions the complexity of the problem increases and so does the training time. To address this issue we consider installing monitoring systems that would serve as reference points for the model and speed up the training process. To find the optimal sensors' placement, we use PINNs and mixed integer optimization. 

This work explores different strategies for finding the optimal number and position of sensors for 1D and 2D spatial problems. The final model not only allows the user to control the spread of the sensors, but it also leverages the distance and transformer temperature to reduce the loss of temperature information that can result from this spread. The proposed model is a general solution to find the optimal sensor placement, regardless of the shape and size of the transformer; therefore, it can be adapted for a variety of applications and potentially used to solve similar problems for other types of electric components that require monitoring.

\section*{Acknowledgment}

This work is supported by the Vinnova Program for Advanced and Innovative Digitalisation (Ref. Num. 2023-00241) and Vinnova Program for Circular and Biobased Economy (Ref. Num. 2021-03748) and partially supported by the Wallenberg AI, Autonomous Systems and Software Program (WASP) funded by the Knut and Alice Wallenberg Foundation.

\bibliographystyle{elsarticle-num} 
\bibliography{references}



\end{document}